\documentclass[11pt,a4paper]{article}
\usepackage[hyperref]{conll-2019}
\usepackage{times}
\usepackage{latexsym}

\usepackage{times}
\usepackage{latexsym}
\usepackage{amsmath}
\usepackage{graphicx}
\usepackage{subfig}
\usepackage[noend]{algpseudocode}
\usepackage{algorithm}

\newcommand{\TA}[1]{\textcolor{magenta}{}}
\newcommand{\MD}[1]{\textcolor{cyan}{}}

\usepackage{url}

\usepackage{comment}
\usepackage{enumitem}
\usepackage{float}

\usepackage{tikz}
\usetikzlibrary{positioning}

\aclfinalcopy 

\title{Global Autoregressive Models for Data-Efficient Sequence Learning}

\author{Tetiana Parshakova \\
  Stanford University\thanks{\ \ Work conducted during an internship at NAVER Labs Europe.} \\
  {\small \tt tetianap@stanford.edu} \\\And
  Jean-Marc Andreoli \\
  Naver Labs Europe \\
  {\hspace{5.5cm}\small \tt \{jean-marc.andreoli,marc.dymetman\}@naverlabs.com } \\
  \And
  Marc Dymetman \\
  Naver Labs Europe \\
  }

\begin{document}
\maketitle
\begin{abstract}
    Standard autoregressive seq2seq models are easily trained by max-likelihood, but tend to show poor results under small-data conditions. We introduce a class of seq2seq models, GAMs (Global Autoregressive Models), which combine an autoregressive component with a log-linear component, allowing the use of global \textit{a priori} features to compensate for lack of data. We train these models in two steps. In the first step, we obtain an \emph{unnormalized} GAM that maximizes the likelihood of the data, but is improper for fast inference or evaluation. In the second step, we use this GAM to train (by distillation) a second autoregressive model that approximates the \emph{normalized} distribution associated with the GAM, and can be used for fast inference and evaluation. Our experiments focus on language modelling under synthetic conditions and show a strong perplexity reduction of using the second autoregressive model over the standard one.
\end{abstract}

\section{Introduction}

Neural sequential text generation models have become the standard in NLP applications such as language modelling, NLG, machine translation. When enough data is available, these models can be trained end-to-end with impressive results. Generally, inference and training proceed in an auto-regressive manner, namely, the next decoded symbol is predicted by a \emph{locally normalized} conditional distribution (the ``softmax''). This has several advantages: (i) the probability of the sequence is already normalized, by the chain-rule over local decisions, (ii) max-likelihood (ML) training is easy, because the log-likelihood of the full sequence is simply the sum of local CE (cross-entropy) losses, (iii) exact sampling of full sequences from the model distribution is directly obtained through a sequence of local sampling decisions.

However, these autoregressive models (AMs) tend to suffer from a form of myopia. They have difficulty accounting for global properties of the predicted sequences, from overlooking certain aspects of the semantic input in NLG 
to duplicating linguistic material or producing ``hallucinations'' in MT, and generally through being unable to account for long-distance consistency requirements that would be obvious for a human reader.\footnote{To borrow terminology from Reinforcement Learning (RL) \cite{Sutton1998}, such NLP models work by ``imitation learning'', without any representation of ``objectives'' to be realized. While this defect can be mitigated in the presence of large training sets, it can become serious when this condition is not met.}

The main contributions of this paper are as follows. 

\emph{First}, we propose a hybrid seq2seq formalization, the \emph{Global Autoregressive Model} (GAM), that combines a local autoregressive component with a global log-linear component, allowing the use of \emph{a priori} features to compensate for the lack of training data. GAMs are related both to the class of Energy-Based Models (EBM) and to that of Exponential Families (EF), and inherit some important properties from those: an intimate relationship between training and sampling (EBM); the identity of empirical and model expectations at maximum-likelihood; convexity of log-likelihood (EF).

\emph{Second}, we propose a training procedure in two steps. In the first step, we train through max-likelihood a GAM, which however is \emph{unnormalized} and improper for fast inference or evaluation. In the second step, we use this GAM to train (by distillation) a second autoregressive model that approximates the \emph{normalized} distribution associated with the GAM, and can be used for fast inference and evaluation. 

\emph{Third}, we demonstrate the ability of GAMs to be data-efficient, namely, to exploit the original data better than a standard autoregressive model. In order to clarify the core techniques and issues, we design a simple class of synthetic data, consisting of random binary strings containing ``motifs'' (specific substrings)  that we can manipulate in different ways. We show that, in limited data conditions, GAMs are able to exploit the features to obtain final autoregressive models that perform better than the original ones.

The remainder of the paper is structured as follows. In Section 2, we provide some background about autoregressive models, energy-based models, and log-linear models. In Section 3, we introduce GAMs. In section 4, we describe our focus on synthetic data. In Section 5, we explain our training procedure. In Section 6, we comment on related work. In Section 7, we describe our experiments. In Section 8, we provide an analysis of our results. We conclude with a discussion in Section 9. \emph{Note that some additional explanations and experiments are provided in the Supplementary Material, indicated by [SM].}

\section{Background}

\subsection{Autoregressive models (AM)} These are currently the standard for neural seq2seq processing, with such representatives as RNN/LSTMs \cite{hochreiter1997long,SutskeverVL14}, ConvS2S \cite{gehring2017convolutional}, Transformer \cite{Vaswani}). Formally, they are defined though a distribution $r_\eta(x|C)$, where $C$ is an input (aka \underline{C}ontext, e.g. a source sentence in Machine Translation (MT)), and $x$ is a target sequence (e.g. a target sentence in MT). We have: 
$$
r_\eta(x|C) \doteq \prod_i s_\eta(x_i|x_1,\ldots,x_{i-1}, C),
$$
where each $s_\eta(x_i|x_1,\ldots,x_{i-1}, C)$ is a normalized conditional probability over the next symbol of the sequence, computed by a neural network (NN) with parameters $\eta$. The local normalization of the incremental probabilities implies the overall normalization of the distribution $r_\eta(x|C)$, and consequently, the possibility of directly sampling from it and evaluating the likelihood of training sequences.

\subsection{Energy-Based Models (EBM)} EBMs are a generic class of models, characterized by an energy function $U_\eta(x|C)$ computed by a NN parametrized by $\eta$ \cite{lecun_tutorial_2006}. Equivalently, they can be seen as directly defining a potential (an unnormalized probability distribution) $P_\eta(x|C) = e^{-U_\eta(x|C)}$, and indirectly the normalized distribution $p_\eta(x|C) = 1/Z_{\eta}(C)\ P_\eta(x|C)$, with $Z_{\eta}(C) = \sum_x P_\eta(x|C)$. A fundamental property of these models is that, for max-likelihood training, the SGD updates can be computed through the formula:\footnote{See \citep[p. 15]{lecun_tutorial_2006}, and [SM] for a derivation.} 
    \begin{align} \label{eq:EBM-SGD}
    \nabla_\eta \log p_\eta(x|C) &= \nabla_\eta \log P_\eta(x|C)\\
    &- E_{x \sim p_\eta(\cdot|C)} \nabla_\eta \log P_\eta(x|C), \nonumber
    \end{align}
which, in principle, reduces the problem of \emph{training} with unnormalized potentials to the problem of \emph{sampling} from them.

\subsection{Log-Linear Models / Exponential Families}

Log-Linear models \cite{Jebara2013} are the conditional version of Exponential Families \cite{Jordan}. The general form of a log-linear model (for the discrete case) is as follows:
    \begin{align*} 
    p_\lambda(x|C) = 1/Z_\lambda(C)\ \mu(x;C)\ e^{\langle\lambda(C),\ \phi(x;C)\rangle},
    \end{align*}
with $Z_\lambda(C) = \sum_x \mu(x;C)\ e^{\langle\lambda(C),\ \phi(x;C)\rangle}$. Here $\phi(x;C)$ is a vector of predefined real features of the pair $(x,C)$, which is combined by scalar product with a real vector of weights $\lambda(C)$ of the same dimension; $\mu(x;C)$ is an arbitrary ``base measure'', which is fixed. These models, which allow to introduce prior knowledge through features and have nice formal properties (see below), were mainstream in NLP before the revival of neural approaches.

\section{Proposal: GAMs} 
We now define \emph{Global Autoregressive Models} (GAMs). These are hybrid seq2seq models that exploit both local autoregressive properties as well as global properties of the full target sequence. A GAM is an unnormalized distribution $P_\eta(x|C)$ over sequences $x$, parametrized by a vector $\eta = \eta_1 \oplus \eta_2$:
\begin{equation}\label{eq:GAM}
P_\eta(x|C) = r_{\eta_1}(x|C) \cdot e^{\langle\lambda_{\eta_2}
(C),\ \phi(x;C)\rangle}.   
\end{equation}
Here $r_{\eta_1}(x|C)$ is an autoregressive seq2seq model for generating $x$ from input $C$, parametrized by $\eta_2$; $\phi(x;C)$ is a vector of predefined real features of the pair $(x,C)$, which is combined by a scalar product with a real vector $\lambda_{\eta_2}(C)$ of the same dimension, computed over the input $C$ by a network parametrized by $\eta_2$. The normalized distribution associated with the GAM is $p_\eta(x|C) = \frac{P_\eta(x|C)}{Z_\eta(C)}$, where $Z_\eta(C) = \sum_x P_\eta(x|C)$.

GAMs appear promising for the following reasons:
\begin{itemize}[leftmargin=*]
    \item Features $\phi(x;C)$ provide a simple way to draw attention of the model to potentially useful aspects that may be difficult for the AM component to discover on its own from limited data. 
    \item GAMs are an instance of EBMs, where the potential $P_\eta(x|C)$ is the product of the an AM potential $r_{\eta_1}(x|C)$ with a ``log-linear'' potential $e^{\langle\lambda_{\eta_2}(C),\phi(x;C)\rangle}$. Here the gradient relative to the log-linear part takes the especially simple form:
    \begin{align} \label{eq:LL-SGD}
    \nabla_{\eta_2} \log p_{\eta}(x|C) &= \phi(x;C)\\ &- E_{x \sim p_\eta(\cdot|C)}\ \phi(x;C). \nonumber
    \end{align}
    \item 
    Log-linear models, on their own, while great at expressing prior knowledge, are not as good as AM models at discovering unforeseen regularities in the data. Also, they are typically problematic to train from a log-likelihood perspective, because sampling from them is often unfeasible. GAMs address the first issue through the $r$ component, and alleviate the second issue by permitting the use of $r$ as a powerful ``proposal'' (aka ``surrogate'') distribution in importance sampling and related approaches, as we will see.
\end{itemize}

\section{Experimental focus \label{p:synth_data}} While the motivation for GAMs ultimately lies in practical NLP applications such as those evoked earlier, in this paper we aim to understand some of their capabilities and training techniques in simple and controllable conditions. We  focus on the unconditional (i.e. language modelling) case, and on synthetic data. Our setup is as follows:
\begin{itemize}[leftmargin=*]
    \item We consider an underlying process $p_{true}$ that generates \emph{binary sequences} according to a well-defined and flexible process. In this paper we use PFSAs (Probabilistic Finite State Automata) to impose the presence or absence of sub-strings (``motifs'') anywhere in the generated data, exploiting the intersection properties of automata. 
    \item Due to the dynamic programming properties of PFSAs, it is possible to compute the true entropy $H(p_{true}) = -\sum_x p_{true}(x) \log p_{true}(x)$ of the process (see [SM]), as well as other quantities (Partition Functions, Mean sequence length); it is also possible to generate training ($D$), validation ($V$), and test data ($T$) in arbitrary quantities.
    \item We employ an unconditional GAM of the simple form:
    \begin{align} 
    p_\lambda(x) &\doteq \frac{P_\lambda(x)}{Z_\lambda}, \text{with } Z_\lambda \doteq \sum_x P_\lambda(x) \text{ and} \nonumber\\
    P_\lambda(x) &\doteq r(x) \cdot e^{\langle \lambda,\ \phi(x)\rangle}, \label{eq:exp-family}
    \end{align}
    where $r$ is trained on $D$ and then kept fixed, and where $\lambda$ is then trained on top of $r$, also on $D$. 
    
    It should be noted that with $r$ fixed in this way, this formulation exactly corresponds to the definition of an exponential family \cite{Jordan}, with $r$ as \emph{base measure}. In such models, we have two important properties: (i) the log-likelihood of the data is \emph{convex} relative to the parameters $\lambda$, and thus a local maximum is also global; (ii) the max-likelihood value $\lambda^{*}$ has the property that the model expectation $E_{x \sim p_{\lambda^{*}}(\cdot)}\ \phi(x)$ is equal to the empirical expectation $|D|^{-1} \sum_{x\in D} \phi(x)$ (``Moment Matching'' property of exponential families).

    \item We are specially interested in the relative \emph{data-efficiency} of the GAM compared to the AM $r$: namely the ability of the GAM to recover a lower perplexity approximation of $p_{true}$ than $r$, especially in small training-set conditions.
\end{itemize}

\section{Training procedure} 
\subsection{Two-stage training}
\begin{figure}[H]
\begin{center}
    \begin{tikzpicture}
        \node (r) at (0,0.5) {$r(x)$};
        \node (pi) at (0,-0.5) {$\pi_\theta(x)$};
        \node (P) at (3,0) {$P_\lambda(x)$};
        \draw[->,>=latex] (r) to[bend left=45] node[below of=mid, pos=0.4, yshift=4mm] {Training-1} (P);
    \draw[->,>=latex] (P) to[bend left=45] node[above of=mid, pos=0.6, yshift=-4mm] {Training-2} (pi);
    \end{tikzpicture}
\end{center}
    \caption{\label{f:tstage} Two-stage training. At the end of the process, we compare the perplexities of $r$ and $\pi_\theta$ on test data: $CE(T,r)$ vs. $CE(T,\pi_\theta)$. \MD{changes to address line 138 of reviews.txt}}
\end{figure}
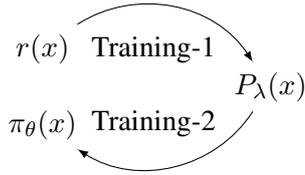

We consider a two-stage training procedure (see Fig.~\ref{f:tstage}).

\paragraph{Training-1} This consists in training the model $P_\lambda$ on $D$. This is done by first training $r$ on $D$ in the standard way (by cross-entropy) and then by training $\lambda$ by SGD with the formula (adapted from (\ref{eq:LL-SGD})): 
\begin{align}  \label{eq:grad_lamda}
\nabla_\lambda \log p_{\lambda}(x) = \phi(x) - E_{x \sim p_\lambda(\cdot)}\ \phi(x).
\end{align}
The main difficulty then consists in computing an estimate of the model moments $E_{x \sim p_\lambda(\cdot)}\ \phi(x)$. In our experiments, we compare two Monte-Carlo approaches \cite{Robert:2005:MCS:1051451} for addressing this problem: (i) \emph{Rejection Sampling} (rs), using $r$ as the proposal distribution and (ii) \emph{Self-Normalized Importance Sampling} (snis) \citep{owen_adaptive_2017,y._bengio_adaptive_2008}, also using $r$ as the proposal.

Rejection sampling is performed as follows. We use $r(x)$ as the proposal, and $P_\lambda(x) = r(x)\ e^{\lambda \cdot \phi(x)}$ as the unnormalized target distribution; for any specific $\lambda$, because our features are bounded between $0$ and $1$, we can easily upper-bound the ratio $\frac{P_\lambda(x)}{r(x)} = e^{\lambda \cdot \phi(x)}$ by a number $\beta$; we then sample $x$ from $r$, compute the ratio $\rho(x) = \frac{P_\lambda(x)}{\beta\ r(x)} \leq 1$, and accept $x$ with probability $\rho(x)$. The accepted samples are unbiased samples from $p_\lambda(x)$ and can be used to estimate model moments.

Snis also uses the proposal distribution $r$, but does not require an upper-bound, and is directly oriented towards the computation of expectations. In this case, we sample a number of points $x_1,\ldots,x_N$ from $r$, compute ``importance ratios'' $w(x_i) = \frac{P_\lambda(x_i)}{r(x_i)}$, and estimate $E_{x \sim p_\lambda(\cdot)}\ \phi(x)$ through $\hat{E} = \frac{\sum_i w(x_i) \phi(x_i)}{\sum_i w(x_i)}$. The estimate is biased for a given $N$, but consistent (that is, it converges to the true $E$ for $N \rightarrow \infty$).

\paragraph{Training-2} While Training-1 results in a well-defined model $P_\lambda(x)$, which may fit the data closely in principle, we should not conclude that $P_\lambda(x)$ is convenient to use for inference --- namely, in language modeling, efficiently sampling from its normalized version $p_\lambda(x)$; as seriously, because of the partition factor $Z_\lambda$, it is also not obvious to \emph{evaluate} the perplexity of $P_\lambda(x)$ on test data.
In order to do both, one approach consists in using a \emph{distillation} technique \cite{DBLP:journals/corr/HintonVD15}, where, during training, one expends generous time towards producing a set of samples from $P_\lambda$, for instance by Monte-Carlo (e.g. Rejection Sampling) techniques, and where this set (which may be arbitrarily larger than the original $D$) is in turn used to train a \underline{new} autoregressive model $\pi_\theta(x)$, which can then be used directly for sampling or for computing data likelihood.
This is the approach that we use in our current experiments, again using the original $r(x)$ as a proposal distribution.
 
\subsection{Cyclical training} In the case of small $|D|$, the proposal distribution $r$ is weak and as a result the distillation process, based on rejection sampling, can be slow. To address this issue, we also consider a cyclical training regime that updates the proposal distribution after distilling each batch of samples, with the intention of reducing the rejection rate. 
Once the process of distillation is finished, we use the aggregated samples to train the final $\pi_\theta$.
The two-stage training procedure is a variant of the cyclical one, with a fixed proposal (see Algorithm~\ref{al:hybrid_algo} for more details).

\begin{algorithm*}
\scriptsize
\caption{Training}\label{al:hybrid_algo}
\begin{algorithmic}[1]
\Function{train}{$D, V, T, ft$, DsSize, tReg, mode}
\State $r \gets \Call{trainRNN}{D,V, \text{optAdam}}$ \Comment{initialize and then train RNN}
\State $P_\lambda \gets \Call{trainGAM}{r, D, V, \text{tReg}, ft}$ \Comment{train $\lambda$ for a given proposal $r$}


\If{mode $=$ `two\_stage'} \Comment{Training-2: distill in one step}
    \State $\widetilde{D}, \ \widetilde{V}, \ \underline{\hspace{2mm}} \gets \Call{distillBatch}{P_\lambda, \text{DsSize}}$ 
\ElsIf{mode $=$ `cyclic'} \Comment{Cyclic-training: distill in several steps} 
\State $\widetilde{D} \gets \{ \}; \widetilde{V} \gets \{ \}; \text{ flag}_{\lambda} \gets \text{False}$
\While{$|\widetilde{D}| < \text{DsSize}$} \Comment{proceed to the distillation process}
    
    \State $\widetilde{D}_B,\ \widetilde{V}_B, \text{ accptRate} \gets \Call{distillBatch}{P_\lambda, \text{bSize}}$ \Comment{accptRate - acceptance rate of $rs$ during distillation}
    
    \State $\widetilde{D}$.insert($\widetilde{D}_B$);  $\widetilde{V}$.insert($\widetilde{V}_B$)
    \If {\textbf{not} $\text{flag}_{\lambda}$}
        \State $r \gets \Call{singleUpdateRNN}{r, \widetilde{D}_B, \text{optAdam}}$ \Comment{improve proposal $r$}
        \State $P_\lambda \gets \Call{trainGAM}{r, D, V, \text{tReg}, ft}$ \Comment{train $\lambda$ for a given proposal $r$}
        \State $\text{flag}_\lambda \gets \Call{earlyStopping$_d$}{\text{acceptRate}}$ \Comment{check if acceptance rate has stopped improving}
        
    \EndIf
\EndWhile
\State $\widetilde{D}$.insert($D$); $\widetilde{V}$.insert($V$) \Comment{add true data to the distilled one}
\EndIf

\State $\pi_\theta \gets \Call{trainRNN}{\widetilde{D}, \widetilde{V}, \text{optAdam}}$
\State \Return $\pi_\theta$
\EndFunction

\Function{trainGAM}{$P_\lambda$, D, V, \text{tReg}, $ft$} \Comment{Training-1}
    \State $\alpha_0 \gets 10$ \Comment{initial learning rate}
    
    \State target$\_mom \gets \Call{getMoments}{D, V, ft}$ \Comment{empirical moments of the given dataset}
    \While{\textbf{not} \Call{earlyStopping}{$\ell_1\_\mathrm{mom}$}} \Comment{check if $\ell_1\_\mathrm{mom}$ has stopped improving}
        \State model\_mom $\gets [0] \times |ft|$ \Comment{accumulate the model's moments} 
        \State $\alpha_t \gets \frac{\alpha_0}{1+\text{\#epoch}}$
        \For{$b \in \text{range}(\#\text{updatesPerEpoch})$}
            \State $\text{mean\_mom} \gets \Call{getMomentsGAM}{P_\lambda, D, V, \text{tReg}, ft}$ \Comment{use $rs$ or $snis$ to estimate $E_{x \sim p_\lambda(\cdot)}\ \phi(x)$}
            \State model\_mom $\gets (\text{model\_mom} + \text{mean\_mom}/(b-1))\cdot \frac{b-1}{b}$ \Comment{moving average}
            \State $\nabla_\lambda \gets  \text{target\_mom} - \text{mean\_mom}$  \Comment{use Eq.~\ref{eq:grad_lamda} to compute gradients}
            \State $\lambda \gets \lambda + \alpha_t \cdot \nabla_\lambda $
        \EndFor
        \State $\ell_1\_\mathrm{mom} \gets \lVert \text{target\_mom} - \text{model\_mom}  \lVert_1$
        
    \EndWhile
    \State \Return $P_\lambda$
\EndFunction

\end{algorithmic}
\end{algorithm*}

\section{Related Work}

\cite{hoang_moment_2018}, working in a NMT context, have a similar motivation to ours. They first train an autoregressive seq2seq model (Transformer in their case) on bilingual data, then attempt to control global properties of the generated sequences through the introduction of \textit{a priori} features. They interpolate the training of the autoregressive model with training of a Moment Matching component which tries to equate the features expectations of the model with those of the data. Contrarily to our approach, they do not directly try to maximize likelihood in an integrated model.

\cite{andor_globally_2016} consider transition-based neural networks, and contrast local to global normalization of decision sequences, showing how the global approach avoids the \emph{label bias} problem in such tasks as tagging or parsing. They focus on inference as maximization, e.g. finding the best sequence of tags for a sequence of words, and consistent with that objective, their training procedure exploits a beam-search approximation. By contrast, our focus is on inference as sampling in a language modelling perspective, on the complementarity between auto-regressive models and log-linear models, and on the relations between training and sampling in energy-based models.


 \begin{figure*}[t!]
    \centering
    \includegraphics[trim = 90 0 0 0, width=1.1\linewidth]{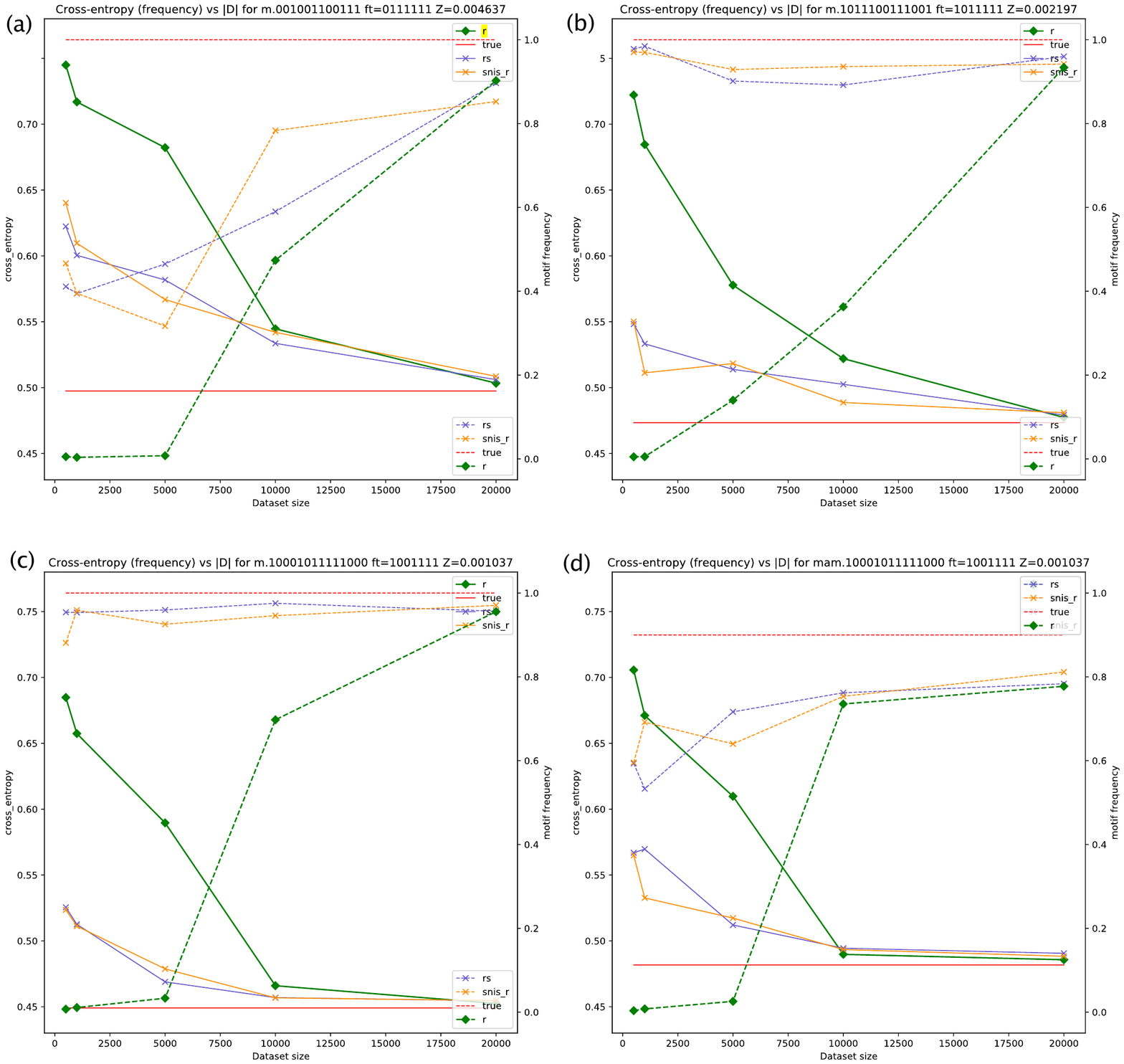} 
    \caption{\label{f:ce_rat} Cross-entropy in nats per character and frequency of sampling motif, depending on $|D|$. Two-stage Training. Features $d_0,\ d_1,\ d_2,\ d_3$ are on for all panels (${ft}_{[4:7]}=\{1111\}$). Panel (a): pure $D$, features $m_{+0}$ (super-motif) and $m_{/2}$ (sub-motif) on; (b): pure $D$, $m$ (motif) and $m_{/2}$ (sub-motif) on; (c) pure $D$, $m$ on; (d) mixture $D$, $m$ on. The plain lines represent cross-entropy, the dashed lines motif frequency.}
\end{figure*}

\section{Experiments}
We conduct a series of experiments on synthetic data to illustrate our approach.
\subsection{Synthetic data}
To assess the impact of GAMs, we focus on distributions $p_{true}(x)$ that are likely to be well approximated by the AM $r(x)$ in the presence of large data. The first class of distributions is obtained through a PFSA that filters binary strings of fixed length $n=30$, $0$'s and $1$'s being equally probable (white-noise strings), through the condition that they contain a specific substring (``motif'') anywhere; here the relative frequency of sequences containing the motif among all sequences varies from $\sim 0.01$ (shorter motifs $|m|=10$) to $\sim 0.001$ (longer motifs $|m|=14$). 
\MD{I added the two parentheses to address a question raised by Reviewer 2 (line 113 of reviews.txt); do you remember what was the range of lengths for motifs?} 

We also consider mixtures of two PFSAs (motif/anti-motif): the first (with mixture prob. $0.9$) produces white-noise strings containing the motif and the second (with mixture prob. $0.1$) strings excluding the motif.

From these processes we produce a training set $D$, of size $|D|$ varying between $5\cdot10^2$ and $2\cdot10^4$, a validation set $V$ of size $0.25 \cdot |D|$ (but never smaller than $5\cdot10^2$ or bigger than $2\cdot10^3$) and a test set $T$ of fixed size $5\cdot10^3$.

\subsection{Features} 
In a real world scenario, prior knowledge about the true process will involve, along with predictive features, a number of noisy and useless features. 
By training the $\lambda$ parameters to match the empirical moments, the GAM will learn to distinguish between these types.
In order to simulate this situation we consider feature vectors over our artificial data that involve both types. 

With $x$ the full string and $m$ the fixed motif used in constructing the training data, we consider variations among the 7 binary features in the set $F$: 
$$
F = \{m,\ m_{+0},\ m_{/2},\ d_0,\ d_1,\ d_2,\ d_3\},
$$
where $m = 0$ iff the motif $m$ appears in $x$, $m_{+0} = 0$ iff the motif followed by a zero (``super-motif'') appears in $x$, $m_{/2} = 0$ iff an initial section of the motif (``sub-motif'', roughly half the size of $m$) appears in $x$. These three features are chosen because they have some correlation with the process for generating the training data. By contrast, the four remaining features are ``distractors'': $d_0 = 0$ iff $x$ begins with a $0$, $d_1 = 0$ (resp. $d_2 = 0$, $d_3 = 0$) iff a certain random, but fixed, string of similar length to $m$ (resp. of larger length, of smaller length) appears in $x$.
We test different configurations of these features for training $\lambda$, and document 
the use/non-use of features with a bit-vector $ft$ of length $|F|$, for instance $ft = 0111111$ means that all features are exploited, apart from $m$.%
\footnote{In the experiments reported here, one of the provided features, $m$, is a detector of the motif actually present in the data generating process, an extreme form of prior knowledge used to illustrate the technique. In general, milder forms of useful prior features can be provided. A simple formal example is to consider one real-valued (non binary) feature for the length, and one for the square of the length, an experiment that we did recently but do not report here; by matching the data expectations of these two additional features, the model is able to represent the mean and variance of length in the data. Here the prior knowledge provided to the model just tells it to be attentive to the distribution of length, a much weaker form of prior knowledge than telling it to be attentive to a specific motif. \MD{an attempt to partially answer line 112 of reviews.txt}}

\subsection{Implementation aspects}
\subsubsection{Autoregressive models} The AMs are implemented in 
PyTorch%
These LSTMs are optimized with Adam \cite{kingma2014adam}, with learning rate $\alpha=0.001$, and with early stopping (patience $=20$) over a validation set. 

\subsubsection{Training: Two-Stage and Cyclical}
The implementation is described in (Algorithm \ref{al:hybrid_algo}). Here we provide some additional details.

\paragraph{Training-1} For training $P_\lambda(x)$ we test two regimes in Eq.~\ref{eq:grad_lamda}, namely $rs$ and $snis$; in both cases, we first train $r(x)$ on the whatever $D$ is available, and use it as the proposal distribution.
During $rs$, we compute the model's expectation over 10 accepted samples, update the $\lambda$'s according to (\ref{eq:grad_lamda}), and iterate. During $snis$, we keep a buffer of the last $5 \cdot 10^4$ samples from $r(x)$ to {compute the weighted average of the feature moments.}
For the training of $\lambda$'s, we use a basic SGD optimization with learning rate $\alpha(\textrm{\#epoch})=\frac{\alpha_0}{1+\textrm{\#epoch}}, \alpha_0=10$. To assess the quality of $P_\lambda(x)$ for early stopping during training, we use the distance between the empirical and model moments:
\begin{equation} \label{eq:l1_mom}
\mkern-10mu \ell_1\_\mathrm{mom} = \bigg \lVert \frac{1}{|D|} \sum_{d \in D} \phi(d) - E_{x \sim p_\lambda(\cdot)}\ \phi(x) \bigg \lVert_1.
\end{equation}

\paragraph{Training-2 and Cyclical Training} When distilling from $P_\lambda$ in Training-2, we use a single proposal $r$, and systematically produce a distilled dataset of size $\text{DsSize}=2\cdot10^4$, which corresponds to the highest value of $|D|$ among those considered for training $r$. In Cyclical Training, the distillation process is performed in several stages, with an evolving $r$ for improving the rejection rate.

\begin{table*}
\begin{center}
\begin{tabular}{lll}
1& $true$          & $101\textbf{10001011111000}1000001001001$                  \\
2&$r$             & $0111110\underline{0001011111000}1110001011$                 \\
3&$\pi_\theta$    & $111010\textbf{10001011111000}0111111100$                 \\
4&$ft$              & ${[}m, \_, \_, d_0, d_1, d_2, d_3{]}$                \\
5&$\lambda$'s     & ${[}-10.1,\_ , \_ , -0.15, -0.06, 0.0, -0.14{]}$ \\
6&mom $true$         & ${[}0.0 ,\_ , \_ , 0.47, 0.99, 1.0, 0.91{]}$    \\
7&mom $r$          & ${[}0.95, \_ , \_ , 0.53, 0.99 , 1.0, 0.91 {]}$ \\
8&mom $\pi_\theta$     & ${[}0.0006, \_ , \_ , 0.43, 0.99, 0.99, 0.91{]}$ \\
9&CEs             & $true$: 0.45, $r$: 0.56, $\pi_\theta$: 0.47                  \\
10&motif freqs     & $true$: 1.0, $r$: 0.045, $\pi_\theta$: 0.959                
\end{tabular}
\end{center}
\vspace{-3mm}
\caption{Illustration. Setting is from Fig. \ref{f:ce_rat}, panel (c):
n =30, motif = 10001011111000 (always present in $D$), ft = 1001111, $|D|$ = 5000, rs used for Training-1.
Lines 1,2,3 show one example from $true,r,\pi_\theta$ respectively; with training set of size 5000, $r$ is only able to generate the motif a fraction of the time (0.045, see line 10), but is better able to generate some submotifs (underlined); $\pi_\theta$ generates the motif frequently (0.959), as illustrated on line 3. With the features from $ft$ (line 4), Training-1 produces a $P_\lambda$ with first feature $\lambda_m$ strongly negative (line 5), meaning that $P_\lambda$ strongly penalizes the absence of the motif; the ``distractor'' features $d_0, d_1, d_2, d_3$ get a weight close to $0$, meaning that they have little predictive power in combination with feature $m$. It is visible from lines 6,7,8 that $\pi_\theta$ is much better able to approximate the true feature expectations than $r$
[features expectations (aka moments) under $r$ (resp. $\pi_\theta$) : $E_{x\sim r(\cdot)}\ \phi(x)$ (resp. $E_{x\sim \pi_\theta(\cdot)}\ \phi(x)$) \MD{added to address line 86 of reviews.txt}]
Finally (line 9), the CE of $\pi_\theta$ relative to the test set is close to the true entropy of the process, while that of $r$ is much further away.}
\label{table:example}
\end{table*}

\begin{table*}[t!]
\begin{center}
\small
\caption{\label{t:reg_time} Comparison of the time for Training-1 in $rs$ and $snis$; for motif $10001011111000$; $ft=1011111$; $H(p_{true})=0.449$ with pure $D$ (m) and $ft=1001111$; $H(p_{true})=0.482$ with mixture of motif-anti-motif $D$ (mam).}
\begin{tabular}{ccccccc}
\hline \rule{0pt}{3ex}$|D|$ & m;$\frac{\text{mtf\_frq}_{rs}}{\text{mtf\_frq}_{snis}}$  & m;$\frac{\textrm{CE}(rs)}{\textrm{CE}(snis)}$& m;$\frac{\textrm{time}(rs)}{\textrm{time}(snis)}$ & mam;$\frac{\text{mtf\_frq}_{rs}}{\text{mtf\_frq}_{snis}}$  & mam;$\frac{\textrm{CE}(rs)}{\textrm{CE}(snis)}$& mam;$\frac{\textrm{time}(rs)}{\textrm{time}(snis)}$  \\ [1mm]  
\hline

\rule{0pt}{3ex}500 & 0.998 & 0.967 & 2.92 & 0.997 & 1.003 & 4.7 \\ [1mm]
\rule{0pt}{3ex}1000 & 1.009 & 0.973 & 2.038 & 0.77 & 1.07 & 3.638 \\ [1mm]
\rule{0pt}{3ex}5000 & 0.995 & 0.967 & 0.756 & 1.12 & 0.99 & 1.365\\ [1mm]
\rule{0pt}{3ex}10000 & 1.134 & 0.956 & 1.514 & 1.011 & 1.002 & 1.005 \\ [1mm]
\rule{0pt}{3ex}20000 & 1.497 & 0.961 & 0.938 & 0.965 & 1.005 & 0.975 \\ [1mm]

\hline
\vspace{-4mm}
\end{tabular}
\end{center}
\end{table*}

\section{Results}

\subsection{Cross-entropy comparison}

We conduct experiments to compare the cross-entropy (measured in nats) between the initial AM $r(x)$ relative to the test set $T$ and the final AM $\pi_\theta(x)$ also relative to $T$; we vary the size of $|D| \in \{ 0.5, 1, 5, 10, 20 \}\cdot10^3$, the regimes (tReg) for Training-1 ($rs$ or $snis$), the features employed, the rarity of the motifs. \figurename~\ref{f:ce_rat} depicts the resulting curves at the end of the two-stage training (plain lines).

Here we show only a few experiments (a more extensive set is provided in the [SM]).

We observe that, for a small dataset size $|D|$, there is a big gap between the CE of $r(x)$ and the CE of $\pi_\theta(x)$. As $|D|$ increases, these cross-entropies become closer to one another, but a large gap persists for $|D| = 5000$.

We note that the presence of the ``fully-predictive'' feature $m$ results in a $\pi_\theta(x)$ that has CE very close to the theoretical entropy, even in low $|D|$ regimes, where $r$ on its own is very weak.\footnote{The CE of a model relative to the true underlying process (approximated by the test set $T$) can never be below the entropy of this process, due to the KL-divergence being non-negative.} Thus, not only is the distilled AM much better than the initial AM, but this is an indication that $P_\lambda$ itself (for which the cross-entropy is more difficult to compute exactly) is a good approximation of the true process.

By contrast, if the $m$ feature is absent, then, while $\pi_\theta$ is still better than $r$ in low $|D|$ regimes, it cannot reach the theoretical entropy in such regimes, because features such as $m_{0+}$ and $m_{/2}$ can only partially model the data. With large $|D|$, on the other hand, $r$ on itself does a good job at predicting the data, and $P_\lambda$ adds little on top of its $r$ component.

Finally, we note that the two regimes for training $P_\lambda(x)$, $rs$ and $snis$, result in $\pi_\theta$'s with similar accuracies. 

We also observe that with a good performance of $\pi_\theta(x)$, the moments of motif feature on the distilled dataset are close to the true ones (see [SM] \figurename~\ref{f:ce_rat_f1}, \ref{f:ce_rat_f2}, \ref{f:ce_rat_mam}).

These trends are consistent across the experiments with different motifs, as can be checked in Table~\ref{t:stats} and with the additional plots in the [SM]. 

\begin{table*}[h!]
\begin{center}
\small
\caption{\label{t:stats} Overall statistics: for $D_m, motif \in \{10001010001,01011101101,001001100111,1011100111001$, \\ \hspace{\textwidth} $10001011111000\}, ft \in \{ 1001111,1011111,0111111\}$ and $D_{mam}$, $motif \in \{01011101101, 001001100111, 1011100111001,$  $100010100011, 10001011111000\}$, $ft \in \{1001111\}$ .}
\begin{tabular}{llccccccc}
\hline \rule{0pt}{3ex}tReg &$|D|$ & m: $\frac{\text{CE}(T, r)}{\text{CE}(T, \pi_\theta)}$ & m: $\frac{\text{CE}(T, \pi_\theta)}{H(p_{true})}$ & m: $\frac{\text{mtf\_frq}(\pi_\theta)}{\text{mtf\_frq}(r)}$ & mam: $\frac{\text{CE}(T, r)}{\text{CE}(T, \pi_\theta)}$ & mam: $\frac{\text{CE}(T, \pi_\theta)}{H(p_{true})}$ & mam: $\frac{\text{mtf\_frq}(\pi_\theta)}{\text{mtf\_frq}(r)}$  \\ [1mm]  
\hline

\rule{0pt}{3ex}$rs$&500 & $1.24\pm0.07$ & $1.19\pm0.07$ & $[32.0,392.0]$ & $1.23\pm0.03$ & $1.16\pm0.03$& $[59.26,433.33]$ \\ [1mm]
\rule{0pt}{3ex}$rs$&1000 & $1.24\pm0.07$ & $1.16\pm0.07$ & $[23.87,653.33]$ & $1.21\pm0.03$ & $1.14\pm0.03$ & $[26.29,233.33]$ \\ [1mm]
\rule{0pt}{3ex}$rs$&5000 & $1.18\pm0.08$ & $1.09\pm0.05$ & $[3.59,206.67]$ & $1.16\pm0.05$ & $1.08\pm0.04$ & $[7.32,130.0]$ \\ [1mm]
\rule{0pt}{3ex}$rs$&10000 & $1.08\pm0.1$ & $1.04\pm0.02$ & $[0.89,196.0]$ & $1.02\pm0.03$ & $1.04\pm0.03$ & $[1.0,4.97]$ \\ [1mm]
\rule{0pt}{3ex}$rs$&20000 & $0.99\pm0.01$ & $1.02\pm0.01$ & $[0.81,1.76]$ & $0.99\pm0.0$ & $1.02\pm0.0$ & $[0.85,1.04]$ \\ [1mm]
\hline
\end{tabular}
\end{center}
\end{table*}

\subsection{Motif frequencies}
In order to assess the \emph{predictive} properties of obtained AMs, we also compare the frequency of motifs in strings sampled from $r$ and from $\pi_\theta$ ($2\cdot10^3$ samples in total).
From \figurename~\ref{f:ce_rat} we see that when vary $|D|$, the frequency of motifs (dashed lines) is aligned with the CE performance. Namely, $\pi_\theta$ produces a higher fraction of strings with motif  than $r$ when $|D|$ is small ($|D|\in \{ 0.5,1,5\}\cdot10^3$).\\[1ex]

\textbf{Detailed illustration} To provide more intuition, we provide an illustration from one experiment in Table \ref{table:example}.

\subsection{Mixture $D_{mam}$ vs pure $D_m$}
In our experiments, the strings in $D_{mam}$ (motif-anti-motif) contain a motif with $p=0.9$. However, if not all of the samples in $D_{mam}$ contain the motif, then the motif feature itself is not fully predictive. It can be seen in panel (d) of  \figurename~\ref{f:ce_rat} that the $\pi_\theta$ achieved with $P_\lambda$ trained on mixture $D_{mam}$ has consistent behaviour with the results obtained on the pure $D_m$ of panels (a,b,c).


\subsection{Regimes in Training-1}
For training GAM we consider two methods, $snis$ and $rs$. As described in the previous sections, their impact on $P_\lambda$ leads to $\pi_\theta$'s that have similar CE's and motif frequencies. Despite such resemblance in terms of accuracy, these two methods differ in terms of speed (see Table~\ref{t:reg_time}). Namely, when $r$ is close to white noise due to small $|D|$, then for the rare events $rs$ rejects most samples not containing the motif due to the effect of the log linear term and negative value of the component  $\lambda_m$ corresponding to the $m$ feature, while $snis$ is able to exploit all samples. Despite being faster than $rs$, $snis$ remains competitive in terms of CE.

\subsection{Cyclical vs two-stage training}
We conducted a small experiment to compare the performance of cyclical training with two-stage training in terms of speed and accuracy for a fixed motif $m$ and features $ft$ (see [SM] Table~\ref{t:cycle_two}, \figurename~\ref{f:cycl_two}). We observed that CEs of the obtained $\pi_\theta$'s were about the same for different values of $|D|$ and Training-1 regimes. On the other hand, there was no systematic improvement in the training speed of one method over the other.


\section{Discussion}

The basic idea behind GAMs is very simple. First, we extend the representational power of the autoregressive model $r$ by multiplying by a log-linear potential, obtaining an unnormalized model $P_\lambda$ (Training-1). Then we try to ``project'' this extended representation again to an autoregressive model $\pi_\theta$ (Training-2). 
Our results showed that, under favorable prior knowledge conditions, the final $\pi_\theta$ was able to perform as well, when trained on small data, as the standard $r$, trained on large data.
During our experiments, we noticed that training $P_\lambda$ was actually easier than training $\pi_\theta$ from it. Intuitively, the small number of parameters to be fitted in the log-linear model requires less work and fewer data than the training of an autoregressive component.\footnote{At a deeper level, there are extreme situations where the $P_\lambda$ obtained at the end of Training-1 can perfectly represent the true process, but where no autoregressive model can actually fit $P_\lambda$: one way to obtain such situations consists in generating binary strings that satisfy a certain cryptographic predicate, associated with a specific feature; the importance of this feature can be easily detected through Training-1, but an autoregressive model has no chance of generalizing from distilled or true data, even in large quantities.}  
 
It is interesting to relate our study to certain aspects of Reinforcement Learning (RL). 

First, consider Training-2. There, we have a ``score'' $P_\lambda$ that we are trying to approximate through an autoregressive model  $\pi_\theta$, which is basically a sequential ``policy''. The main difference with RL is that we are not trying to find a policy that \emph{maximizes} the score (which would be a bad idea for language modelling, as it would tend to concentrate the mass on a few sequences), but one that approximates $P_\lambda$ in a \emph{distributional} sense; our current distillation technique is only one way to approach this problem, but other techniques more in the spirit of RL are possible, a direction that we leave for future work.

Second, consider Training-1. Our approach, consisting in suggesting to the model a number of prior features, might look too easy and suspicious. But notice that in RL, one would typically directly provide to the model an externally defined \emph{reward}, a very strong form of prior knowledge. Here, instead, we ``only'' indicate to the models which features it might attend to, and Training-1 then determines the ``reward'' $P_\lambda$ through max-likelihood, a milder form of prior knowledge, more respectful for what the data has to say.\footnote{We could say that while Training-2 addresses a question directly related to Reinforcement Learning, Training-1 addresses one related to \emph{Inverse} Reinforcement Learning \cite{Russell:1998:LAU:279943.279964,Ng:2000:AIR:645529.657801}: it derives a reward from training evidence rather than imposing it externally.}

\vspace{\fill}
{\small
\subsubsection*{Acknowledgements}
Thanks to Matthias Gall\'e and Ioan Calapodescu for comments on a previous version of this paper and to the anonymous reviewers for their detailed reading and feedback.
}
\vspace{1.5cm}


\clearpage

\clearpage
\appendix
\section{Supplementary Material}

\subsection{SGD in Energy-based models}

The formula \ref{eq:EBM-SGD} is fundamental for studying the SGD behavior of Energy-based models, and for convenience, we provide a derivation here.

If we define $Z_\eta(C) \doteq \sum_x P_\eta(x|C)$, we find that:
\begin{align*}
    \nabla_\eta \log Z_\eta(C) &= \nabla_\eta \log \sum_x P_\eta(x|C)\\
    &\hspace{-2em}= \frac{1}{Z_\eta(C)} \nabla_\eta \sum_x P_\eta(x|C)\\
    &\hspace{-2em}= \frac{1}{Z_\eta(C)} \sum_x \nabla_\eta P_\eta(x|C)\\
    &\hspace{-2em}= \frac{1}{Z_\eta(C)} \sum_x P_\eta(x|C) \nabla_\eta \log P_\eta(x|C)\\
    &\hspace{-2em}= \sum_x p_\eta(x|C) \nabla_\eta \log P_\eta(x|C)\\
    &\hspace{-2em}= E_{x \sim p_\eta(\cdot|C)} \nabla_\eta \log P_\eta(x|C).\\
\end{align*}
We then have: 
\begin{align*}
    \nabla_\eta \log p_\eta(x|C) &= \nabla_\eta \log P_\eta(x|C)- \nabla_\eta \log Z_\eta(C)\\
    &= \nabla_\eta \log P_\eta(x|C)\\
    &\qquad - E_{x \sim p_\eta(\cdot|C)} \nabla_\eta \log P_\eta(x|C).\\
\end{align*}

\subsection{The relevance of finite state automata; connections and differences with Reinforcement Learning \label{a:pfsa}}

The way our synthetic data is produced through FSA's may look contrived, but there are good motivations for using automata in such a study as ours.

Consider the following problem: you are given some RNN $r$ that produces sequences $x$ over a vocabulary $V$, with probabilities $r$ but you would like to filter out sequences that \emph{do not} contain a specific symbol $a$, while preserving the relative probabilities of sequences provided by the RNN: $p_{filtered}(x) \propto P_{filtered}(x) = r(x)\ I[{a \in x}]$. There appears to be no obvious way to realize $p_{filtered}$ through an RNN, apart from techniques similar to what we have been describing in our discussion of Training-2.

The situation is completely different with FSA's. If you have a PFSA (Probabilistic FSA) $r_{pfsa}$ generating sequences $x$, then you can intersect $r_{pfsa}$ with an automaton that accepts all sequences containing at least one $a$, and re-normalize the intersection through dynamic programming, obtaining a new PFSA that generates the filtered distribution.\footnote{And this is exactly what we do to produce training data for our experiments, but using a binary sequence (motif $m$) instead of a single symbol $a$.} Such dynamic programming, with the capacity to \emph{anticipate} properties that need to be satisfied on the global sequence, is unavailable in the RNN world.

With RNNs, the situation is reminiscent of RL, with a reward associated with having observed an $a$ during the production of the sequence. But a standard RL approach would mean that we would try to \emph{maximize} $P_{filtered}(x)$, without taking into consideration the original $r(x)$ that we are filtering from. To be correct, we need to find a policy $\pi_\theta(x)$ (similar to an RNN), that tries to \emph{approximate} $p_{filtered}(x)$ in a \emph{distributional} sense, not in a \emph{maximization} sense (see \citep{bellemare_distributional_2017} for related considerations). This is what we try to do in Training-2, using motifs as our main case-study, instead of a single symbol $a$ (which would not make sense for binary strings).

The advantages of using PFSAs in our study are multiple. They provide a well-understood comparison point to the more complex techniques that need to be deployed for autoregressive models. From an operational viewpoint, they also permit, through dynamic programming, to perform various calculations of interest for our study, such as sampling datasets of arbitrary size and computing exact entropy and partition functions that can serve as comparison points for the results obtained with GAMs. In the present paper, we only exploited PFSA's in the context of motifs, but they provide a much larger class of models that could serve to expand our understanding of sequence-based energy based models.

\subsubsection{Computing the Entropy of a PFSA}

As mentioned earlier, one advantage of using weigthed finite-state automata for generating synthetic data is that some important quantities, such as entropy, mean sequence length, or partition function can be computed by dynamic programming. 

Here we only derive a simple iterative method for computing the entropy of a PFSA, the other computations are very similar.\footnote{For another technique, and for extensions to the computation of relative entropy, see \cite{Carrasco97accuratecomputation}.}

We consider a PFSA with transitions of the form $(q,l,q',w)$, where $q,q'$ are states, $l$ is the label of the transition from $q$ to $q'$ (in our case $l \in \{0,1\}$), and $w$ is the probability of the transition. The fact that the automaton is \emph{probabilistic}, instead of simply \emph{weighted}, means that the sum of $w$'s associated with transitions starting at $q$ is equal to $1$. We further assume that the automaton is \emph{deterministic}, namely that given $q$ and $l$ uniquely determines the next state $q'$.\footnote{The case of non-deterministic probabilistic automata appears much more difficult \cite{CortesMRR08}.}

The entropy $H(q)$ of a state $q$ is defined as $H(q) \doteq -\sum_{x\succ q} p(x|q) \log p(x|q)$, where $x\succ q$ denotes a sequence of labels $x$ that ends in a final state of the automaton, for which $p(x|q)$ is computed in the obvious way. The entropy of the automaton as a whole is then defined as $H(q_s)$, where $q_s$ is the initial state of the automaton.

\textbf{Lemma} \textit{The entropies of states satisfy the fixpoint equation:}
\begin{equation}\label{eq:entropy_fixpoint}
    H(q) = \sum_{(q,l,q',w)} - w \log w + w H(q').
\end{equation}

\noindent Proof. Let's denote by $q_l$ the state obtained from $q$ by following label $l$. We have:
\begin{align*}
    H(q) 
        &= -\sum_{x\succ q} p(x|q) \log p(x|q)\\
        &= -\sum_{{l \cdot y} \succ q} p({l \cdot y}|q) \log p({l \cdot y}|q)\\
        &= -\sum_{{l \cdot y} \succ q} p(l|q) p(y|q_l) [\log p(l|q) + \log p(y|q_l)]\\
        &= -\sum_{l} p(l|q) \log p(l|q) \sum_{y\succ q_l} p(y|q_l) \\
          &\quad -\sum_{l} p(l|q) \sum_{y\succ q_l} p(y|q_l) \log p(y|q_l) \\
        &= -\sum_{l} p(l|q) \log p(l|q) + \sum_{l} p(l|q) H(q_l) \\
        &= \sum_{(q,l,q',w)} - w \log w + w H(q').\\
\end{align*}

It is  possible to show that the state entropies actually correspond to the \emph{least} fixpoint of equation (\ref{eq:entropy_fixpoint}), and this allows a simple iterative algorithm for computing the state entropies: at time $t=0$, for all states $q$, we define $H^{t=0}(q) \doteq 0$, and then we iterate until convergence:
$$
H^{t+1}(q) = \sum_{(q,l,q',w)} - w \log w + w H^{t}(q').
$$

\subsection{Additional Experiments and Results}
(See next pages)


\begin{table*}
\begin{center}
\small
\caption{\label{t:cycle_two} Cyclical training vs two stage training for motif $10001011111000, D_m, ft=1001111$; $\text{CE} \text{ is short for } \text{CE}(T, \pi_\theta).$}
\begin{tabular}{ccccc}
\hline \rule{0pt}{3ex}$|D|$ & $\frac{\text{CE}_{rs}}{\text{CE}_{rs}(\text{cycl})}$ & $\frac{\text{time}_{rs}}{\text{time}_{rs}(\text{cycl})}$ & $\frac{\text{CE}_{snis}}{\text{CE}_{snis}(\text{cycl})}$ & $\frac{\text{time}_{snis}}{\text{time}_{snis}(\text{cycl})}$ \\ [1mm]  
\hline

\rule{0pt}{3ex}500 & 1.02 & 1.21 & 1.02 & 1.51  \\ [1mm]
\rule{0pt}{3ex}1000 & 1.0 & 1.48 & 1.08 & 2.04  \\ [1mm]
\rule{0pt}{3ex}5000 & 1.04 & 0.57 & 1.0 & 0.57  \\ [1mm]
\rule{0pt}{3ex}10000 & 0.98 & 1.47 & 1.02 & 0.45 \\ [1mm]
\rule{0pt}{3ex}20000 & 0.99 & 2.65 & 1.0 & 0.28  \\ [1mm]

\hline
\end{tabular}
\end{center}
\end{table*}

 \begin{figure*}[t!]
    \centering
    \includegraphics[width=0.85\linewidth]{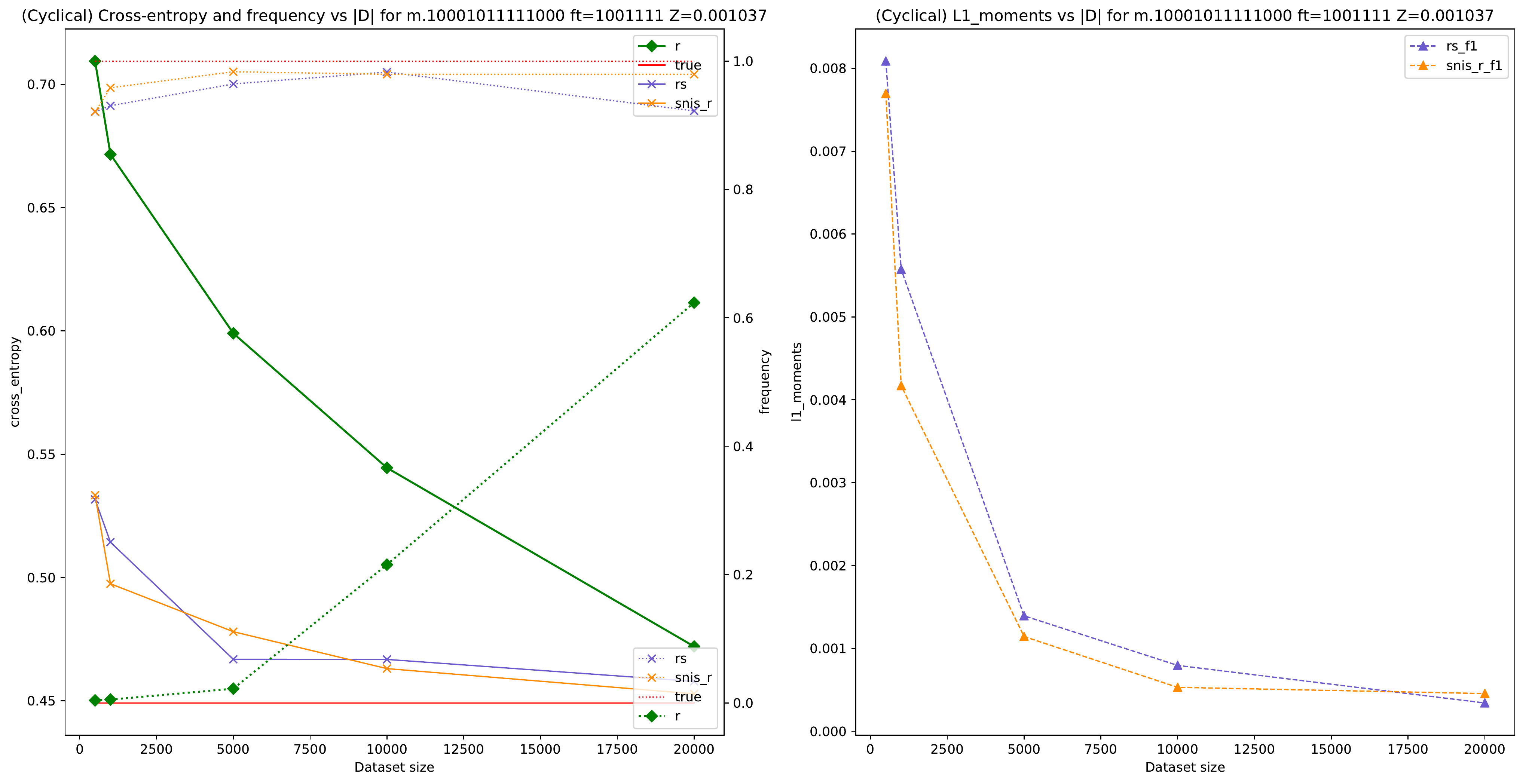} 
    \caption{\label{f:cycl_two} (Cyclical training) Cross-entropy in nats per character, $\ell_1\_\text{mom}$ and frequency of sampling motif, depending on the $|D|$, while all distractive features and motif feature are 1: ${ft}_{[4:7]}=\{1111\}, \ {ft}_{[1]}=1.$  }
\end{figure*}

\begin{figure*}[t]
    \centering
    \includegraphics[width=0.95\linewidth]{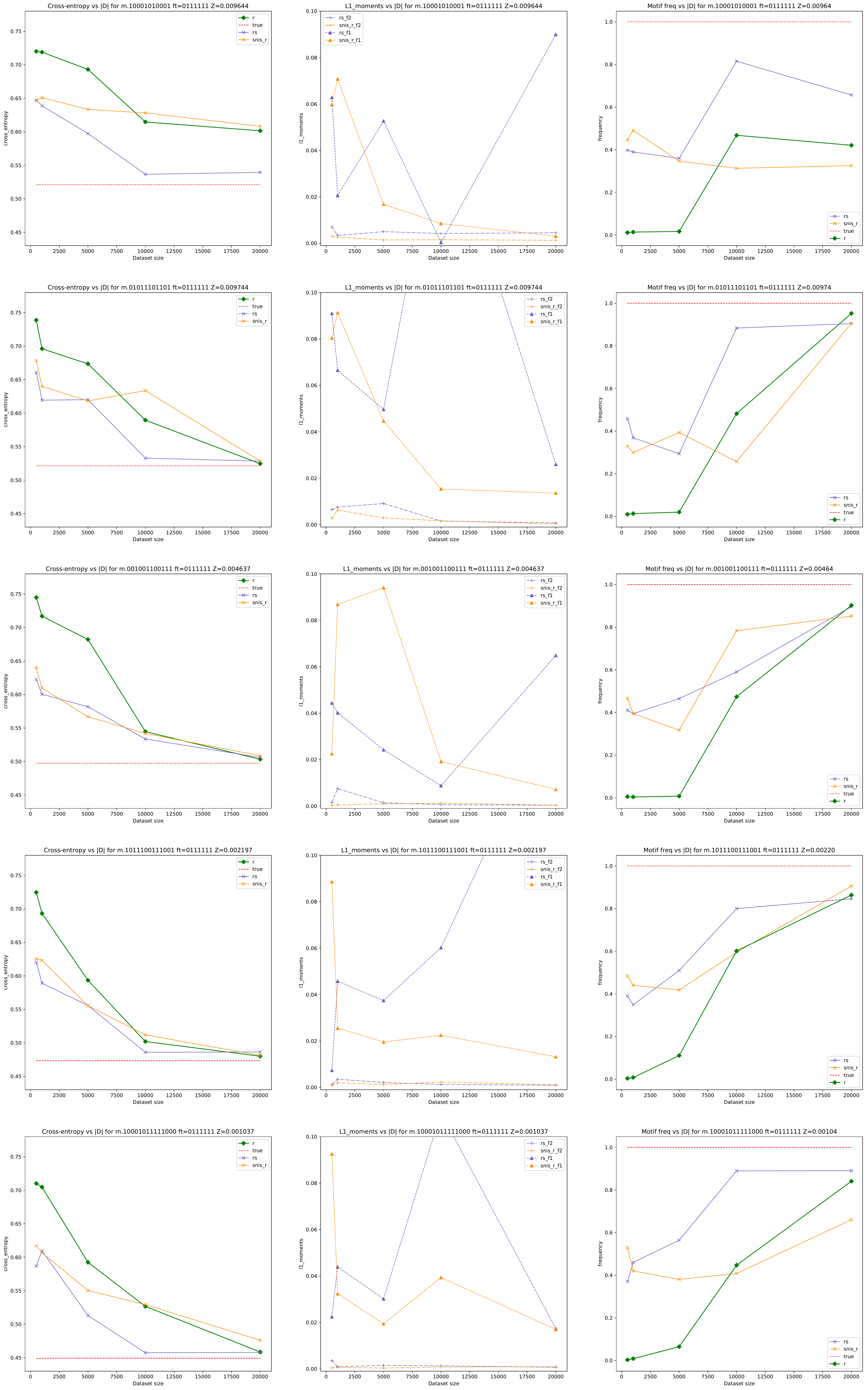} 
    \caption{\label{f:ce_rat_f1} Column 1: Cross-entropy; column 2: $\ell_1\_\mathrm{mom}$; column 3: frequency of sampling motif, depending on the $|D|$, all distractive features are 1: ${ft}_{[4:7]}=\{1111\}$. Setting: supermotif+submotif, pure $D$, while varying the rareness of the motif ($Z$). }
\end{figure*}

\begin{figure*}[t]
    \centering
    \includegraphics[width=0.95\linewidth]{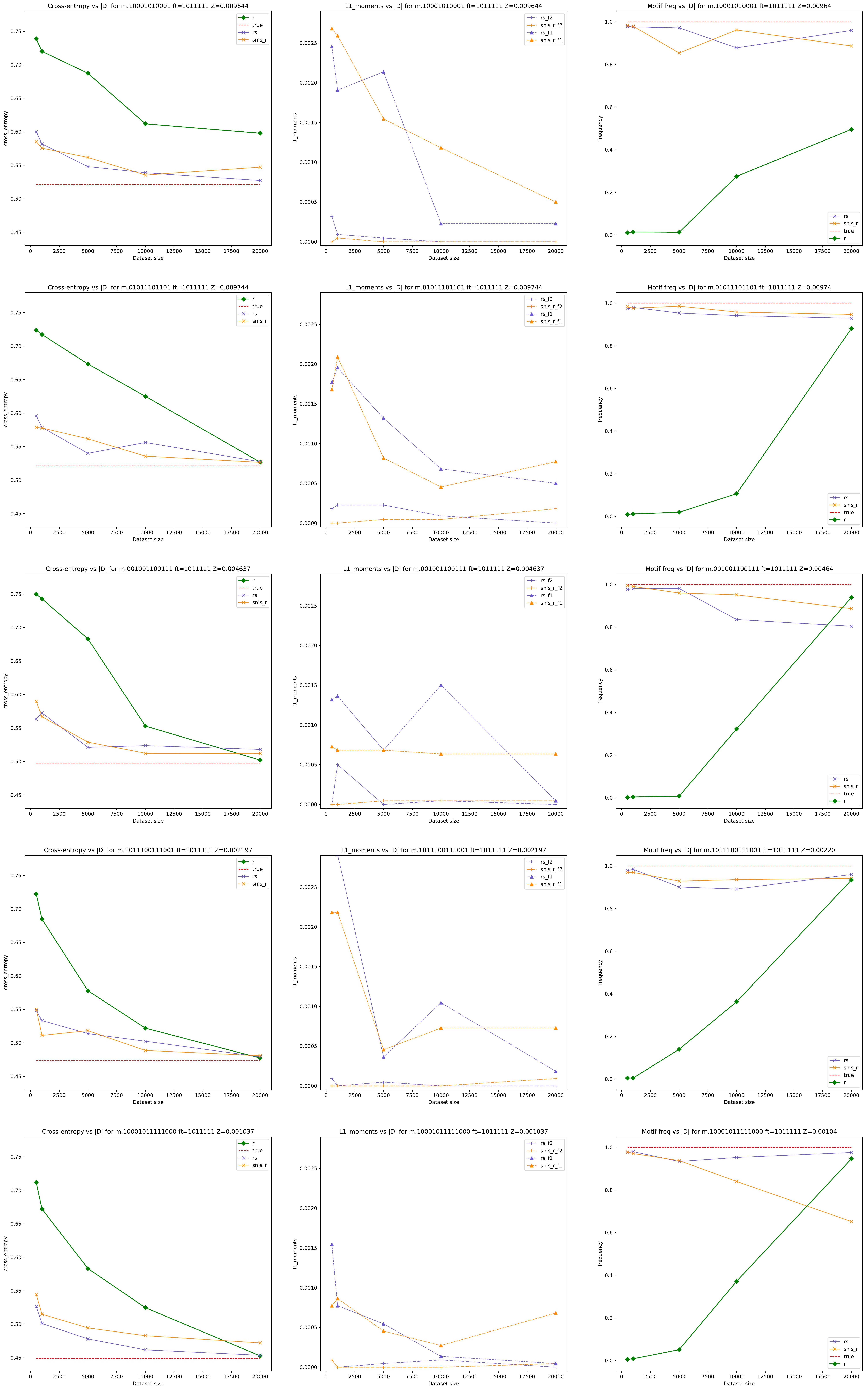} 
    \caption{\label{f:ce_rat_f2} Column 1: Cross-entropy; column 2: $\ell_1\_\mathrm{mom}$; column 3: frequency of sampling motif, depending on the $|D|$, all distractive features are 1: ${ft}_{[4:7]}=\{1111\}$.  Setting: motif+submotif, pure $D$, while varying the rareness of the motif ($Z$).}
\end{figure*}

\begin{figure*}[t]
    \centering
    \includegraphics[width=0.95\linewidth]{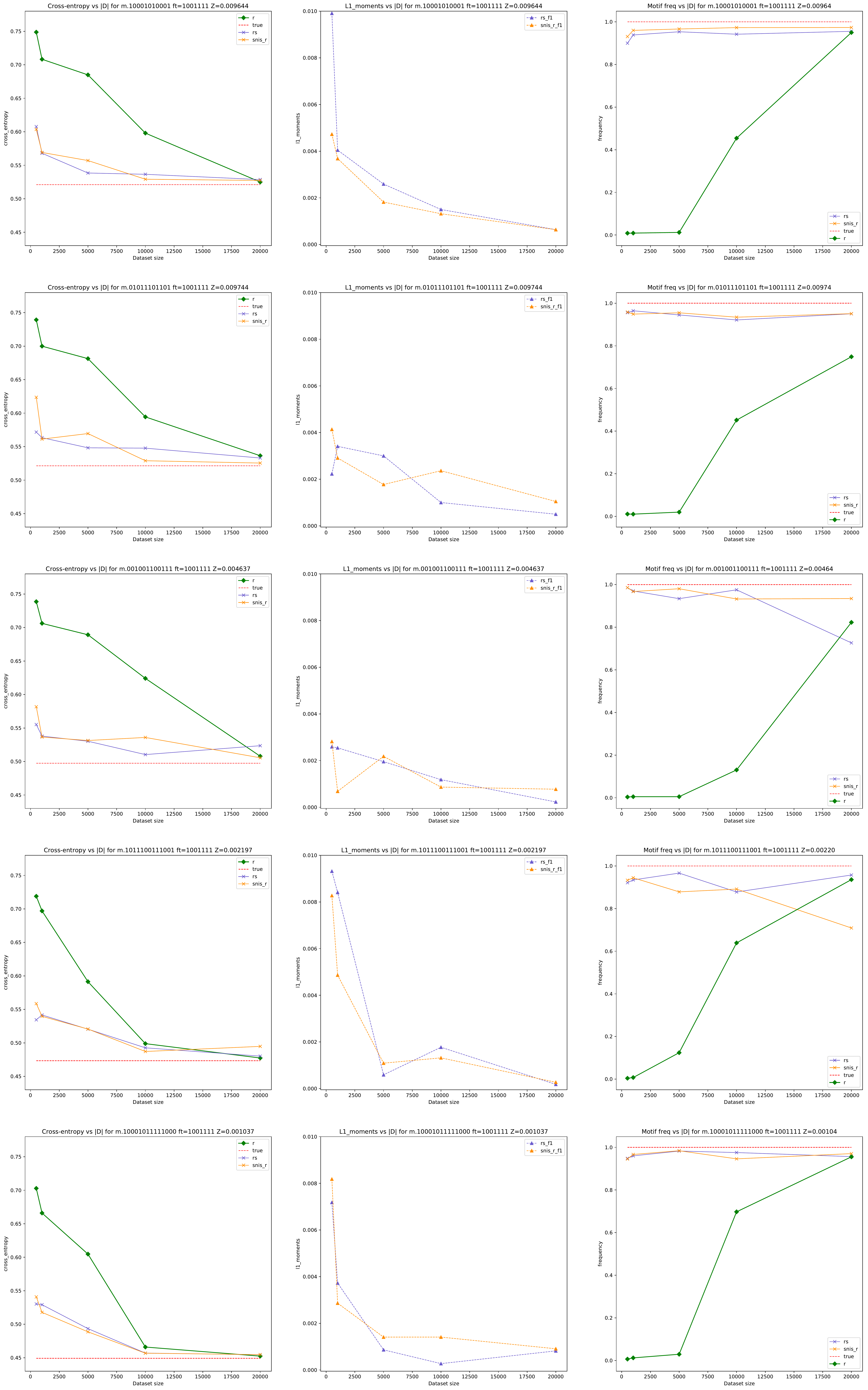} 
    \caption{\label{f:ce_rat_f3} Column 1: Cross-entropy; column 2: $\ell_1\_\mathrm{mom}$; column 3: frequency of sampling motif, depending on the $|D|$, all distractive features are 1: ${ft}_{[4:7]}=\{1111\}$.  Setting: motif, pure $D$, while varying the rareness of the motif ($Z$).}
\end{figure*}

\begin{figure*}[t]
    \centering
    \includegraphics[width=0.95\linewidth]{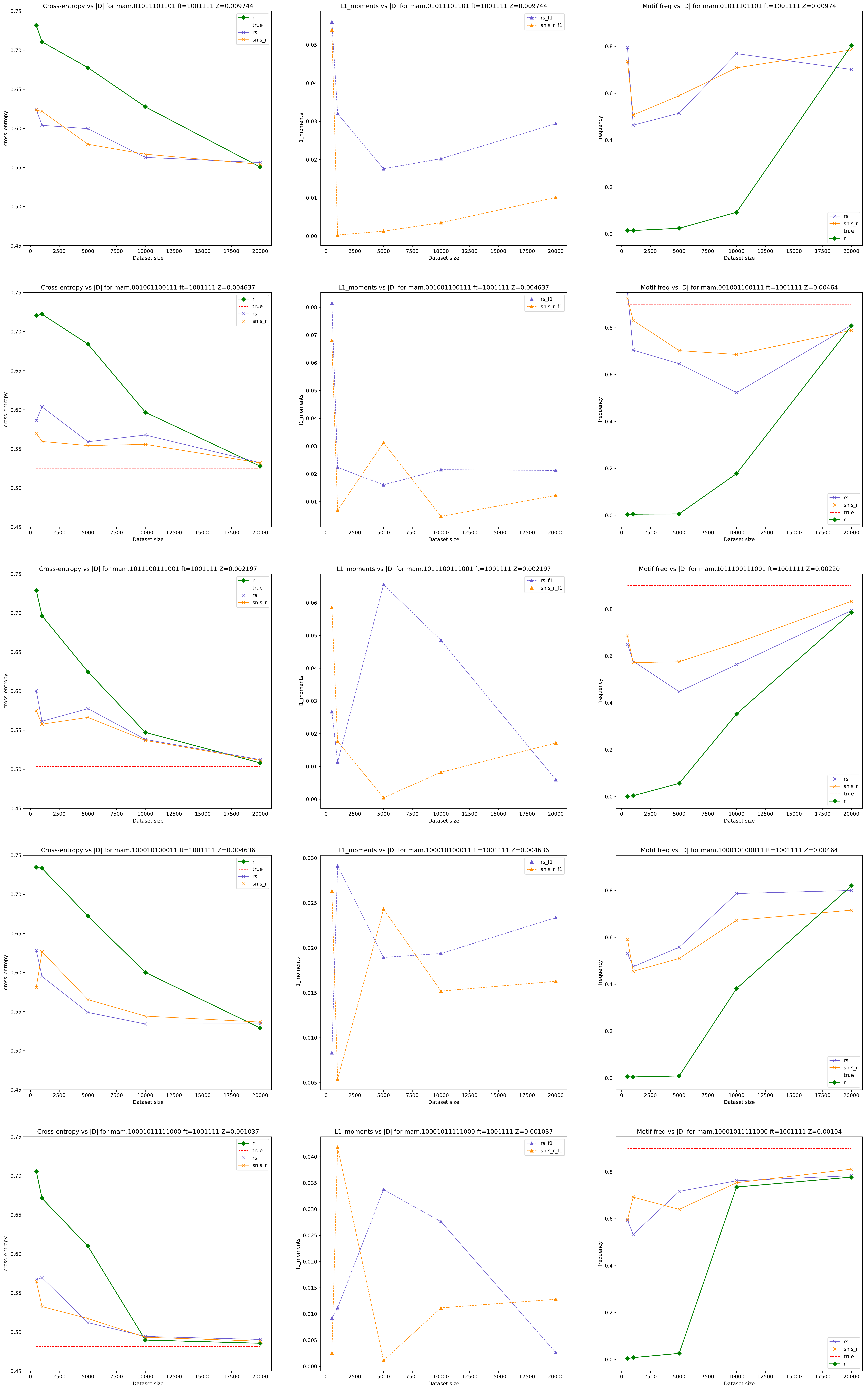} 
    \caption{\label{f:ce_rat_mam} Column 1: Cross-entropy; column 2: $\ell_1\_\mathrm{mom}$; column 3: frequency of sampling motif, depending on the $|D|$, all distractive features are 1: ${ft}_{[4:7]}=\{1111\}$.  Setting: motif, mixture $D$, while varying the rareness of the motif ($Z$).}
\end{figure*}

\end{document}